\documentclass[11pt,a4paper]{article}
\usepackage{authblk}
\usepackage[hyperref]{acl2021}
\usepackage{times}
\usepackage{graphicx}
\usepackage{caption}
\usepackage{latexsym}

\usepackage{multirow}
\usepackage{hyperref}

\usepackage{microtype}

\aclfinalcopy 
\def\aclpaperid{127} 

\title{More Parameters? No Thanks!}

\author[1]{\textbf{Zeeshan Khan}}     
\author[1]{\textbf{Kartheek Akella}}     
\author[2]{\textbf{Vinay P. Namboodiri}}   
\author[1]{\textbf{C V Jawahar}}
\affil[1]{CVIT, IIIT-H}
\affil[2]{University of Bath}
\affil[ ]{\texttt {\{zeeshank606,sukruthkartheek\}@gmail.com}}
\affil[ ]{\texttt {vpn22@bath.ac.uk, jawahar@iiit.ac.in}}
\date{}

\begin{document}
\maketitle

\begin{abstract}
This work studies the long-standing problems of model capacity and negative interference in multilingual neural machine translation ({\sc mnmt}). We use network pruning techniques and observe that pruning 50-70\% of the parameters from a trained {\sc mnmt} model results only in a 0.29-1.98 drop in the {\sc bleu} score.  Suggesting that there exist large redundancies even in {\sc mnmt} models. These observations motivate us to use the redundant parameters and counter the interference problem efficiently. We propose a novel adaptation strategy, where we iteratively prune and retrain the redundant parameters of an {\sc mnmt} to improve bilingual representations while retaining the multilinguality. Negative interference severely affects high resource languages, and our method alleviates it without any additional adapter modules. Hence, we call it parameter-free adaptation strategy, paving way for the efficient adaptation of {\sc mnmt}. We demonstrate the effectiveness of our method on a 9 language {\sc mnmt} trained on {\sc ted} talks, and report an average improvement of +1.36 {\sc bleu} on high resource pairs. Code will be released \href{https://github.com/zeecoder606/PF-Adaptation}{here}.
\end{abstract}

\section{Introduction}

Multilingual neural machine translation({\sc mnmt}) has seen various advances in recent years \citep{dong-etal-2015-multi, firat-etal-2016-multi, zoph-etal-2016-transfer, tan2018multilingual, aharoni-etal-2019-massively, arivazhagan2019massively}. However, the core principle behind the effectiveness in terms of modelling multiple languages remains the same, i.e., sharing all the model parameters between all the languages \citep{johnson-etal-2017-googles}. Although highly scalable and effective, the performance on high resource languages decreases as more low resource languages are added in the model; this is called negative interference. To overcome this, recent works \citep{bapna-firat-2019-simple, philip-etal-2020-monolingual, zhang-etal-2020-improving} proposed language-specific adapter modules, which provide extra parameters to learn language specific representations, and overcomes the effect of negative interference caused by a high degree of parameter sharing.

\begin{figure*}
  \centering
  \includegraphics[width=6.2in]{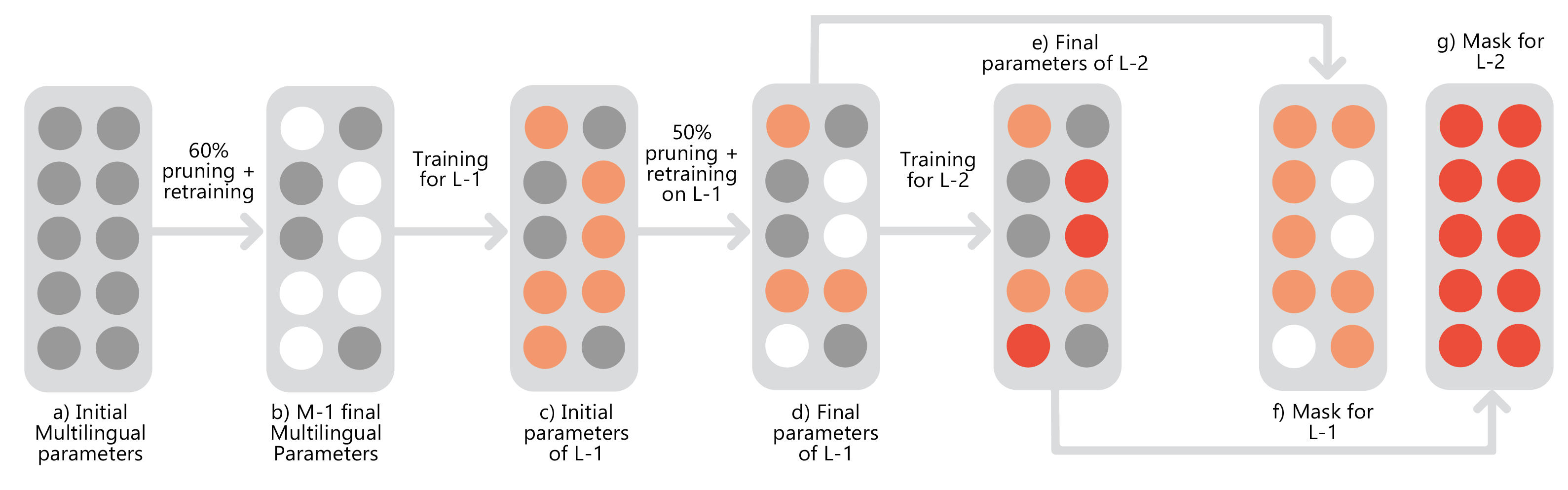}
  \caption{(Better seen in colour.) Illustration of the evolution of model parameters. (a) shows the multilingual parameters in grey. Through 60\% pruning and retraining, we arrive at (b), here white represents the free weights with value=0. The surviving weights in grey will be fixed for the rest of the method. Now, we train the free parameters on the first bilingual pair (L-1) and arrive at (c), which represents the initial parameters of L-1 in orange, and share weights with the previously trained multilingual parameters in grey. Again, with 50\% pruning and retraining on the current L-1 specific weights in orange, we get the final parameters for L-1 shown in (d) and extract the final mask for L-1 in (f). We repeat the same procedure for all the bilingual pairs and extract the masks for each pair.}
  \label{training}
\end{figure*}

In this paper, we propose an alternative to adapter modules. Instead of adding more parameters, we show that the Transformer \citep{NIPS2017_3f5ee243} has enough capacity to model multiple languages and overcome negative interference effectively. 
Inspired by the work of \citet{mallya2018packnet}, we apply iterative pruning to free up the redundant parameters from an {\sc mnmt}, and retrain them to learn language specific representations. We start with a trained {\sc mnmt} model, and prune a fraction of the model parameters, we freeze the surviving parameters and retrain the free ones on a bilingual dataset. This process is iteratively applied for each bilingual pair to get bilingual masks over all the model parameters, as illustrated in figure \ref{training}. We show that using only a fraction of redundant parameters, significantly improves the performance on high resource languages. Also, we retain the multilinguality and the zero-shot translation ability after adaptation. By demonstrating the effectiveness of this approach, we open a potential research direction towards parameter-free adaptation in {\sc mnmt}.

\section{Related Work}
\textbf{Adding multiple tasks to a single network:} Due to the over-parameterized nature of deep neural networks, prior works \citep{Kirkpatrick3521, NIPS2017_f708f064, li2017learning, 8237410} aimed at developing methods to learn multiple tasks while avoiding catastrophic forgetting. \citet{mallya2018packnet} proposed an iterative pruning approach to free up parameters for adding new tasks and retain the previously trained parameters at the same time. Inspired by the concept, we show that an {\sc mnmt} Transformer model can be heavily pruned and the freed up parameters can be retrained to improve bilingual performance, while retaining the multilinguality.

\textbf{Adapting multilingual model to a new language pair and domain adaptation:} Prior works on adaptation 
\citep{neubig-hu-2018-rapid, varis-bojar-2019-unsupervised, stickland2020recipes, escolano2020multilingual, akella2020exploring, bapna-firat-2019-simple, philip-etal-2020-monolingual, zhang-etal-2020-improving} aims at improving language specific performance by either fine-tuning the same {\sc mnmt} model or adding language specific modules. While being effective, these methods either lose their multilinguality or introduce additional parameters. Sharing the same objective, we propose a method to adapt an {\sc mnmt}, without adding language-specific modules, while retaining the multilinguality at the same time. Another line of work \citep{thompson-etal-2018-freezing, wuebker-etal-2018-compact}, proposed training of subnetworks and freezing the rest for domain adaptation. 

\section{Method}
The central idea of our method is to use magnitude pruning to free up parameters in the model and learn bilingual specific representations. Figure \ref{training} depicts the evolution of model weights during the training procedure, with (a) representing the initial multilingual weights in grey. We prune away a fraction of parameters using the one-shot magnitude pruning technique \citep{NIPS2015_ae0eb3ee}, which results in a compressed multilingual representation.
We further train the survived multilingual weights for a few more epochs on the multilingual dataset to compensate for extreme pruning, now the multilingual parameters will remain fixed. Then, we use the free parameters to learn the first language-specific representations. 
We select the first bilingual dataset and train the free parameters. Next we again prune a fraction of weights from the current  bilingual parameters only, to accommodate more bilingual representations. We repeat the same procedure for all the existing bilingual pairs.
A point to note is that during a forward pass data flows through all the shared and specific weights, while during the backward pass only the current bilingual-specific parameters get updated. Hence, the accuracy is retained for all the previously trained bilingual pairs and it enables a high degree of sharing and specificity at the same time. 

\textbf{Pruning Approach:} We perform magnitude pruning \citep{NIPS2015_ae0eb3ee} over the weights of all layers. For simplicity, we do not use the more sophisticated pruning methods \citep{LotteryTicketFrankleC19, NEURIPS2019_2c601ad9, voita-etal-2019-analyzing}. We do not perform pruning over biases and layer normalization parameters, since they correspond to less than 1\% of the total parameters, which is insignificant. Also, we do not prune the embeddings, as they are data specific parameters. All are kept fixed after training the multilingual model. 

\textbf{Inference:}
After finishing the training for each bilingual pair, we get the final mask over all the parameters of the model.  Values of the mask range from $1\rightarrow N$, where $N$ is the total number of bilingual pairs. Each model parameter is masked according to the bilingual pair of interest. To predict a translation for the $t^{th}$ pair, all the parameters learned for languages $1\rightarrow t$ will be used, as shown in figure \ref{training}(f) and (g).

\section{Experiments}

\subsection{Datasets}
We use the {\sc ted} talks \citep{qi-etal-2018-pre} in all our experiments, and all the numbers are {\sc bleu} \citep{papineni-etal-2002-bleu} scores over the test set\footnote{Scores reported are SacreBLEU \citep{post-2018-call}}. Here we have chosen to train on 8 English centring language pairs\footnote{\texttt{ar, az, be, de, gl, he, it, sk}} en-xx covering a spectrum of sizes from high resource \emph{Ar (Arabic)}, 214K to low resource \emph{Be (Belarusian)}, 4.5K.

\subsection{Training}
\textbf{Architecture:} We use Transformer architecture \citep{NIPS2017_3f5ee243}, implemented in \texttt{fairseq} \citep{ott-etal-2019-fairseq}, which was modified to include the pruning and masking modules. We train a joint {\sc bpe} model \citep{sennrich-etal-2016-neural} on all languages to the vocabulary size of 40K. The Transformer \citep{NIPS2017_3f5ee243} architecture used in this work\footnote{\texttt{transformer in fairseq}} has 8 attention heads, 6 encoder and decoder layers, an embedding size of 512, and a feed-forward dimension of 2048. We set the dropout to 0.3.

\textbf{MNMT Training:}
We train a standard {\sc mnmt} model following similar settings as \citet{johnson-etal-2017-googles}. A single many-to-many model is trained on all the English-centric data, using a source-side control token to indicate the target language. We use Adam \citep{DBLP:journals/corr/KingmaB14} with an inverse square root schedule, with 4500 warm-up updates and a maximum learning rate of 0.0003. We set the maximum batch size per {\sc GPU} to 3050 tokens and train on 4 GPUs. Like \citet{arivazhagan2019massively}, to avoid the size imbalance, we use the temperature-based sampling strategy with $T=5$. The {\sc mnmt} is trained for 40 epochs over 8 English-centric language pairs, i.e., 16 directions. As shown in table \ref{table:results}, we train a strong parent {\sc mnmt} baseline.

\textbf{Pruning MNMT:} We prune 50\% of parameters from a fully converged {\sc mnmt} model, and retrain the surviving parameters on the same multilingual dataset for ten more epochs, to compensate for the lost parameters. 

\textbf{Adapting MNMT to bilingual specific representations:}
After pruning the {\sc mnmt} model, we select each bi-direction datasets (en-xx and xx-en) in the descending order of dataset sizes. We use the original source side control token, reset the learning rate scheduler and train all the free parameters for 20 epochs. Then, we prune 75\% of parameters from the current bilingual specific parameters and retrain for ten more epochs to compensate for heavy pruning. 

Pruning ratios are decided based on the trade off between the accuracy lost and the space left to adapt all the languages. We prune 50-70\% of parameters from the parent {\sc mnmt} and observe that it leads to a drop of 0.29-1.98 Bleu score. Therefore, we select 50\% to be the first pruning ratio, and is kept constant in all the experiments. The second pruning ratio is kept 75\% such that the last language pairs get at least 2-5\% of parameters. More variations in the second pruning ratio is demonstrated in section 5.4.

\begin{table*}
 \centering
 \begin{tabular}{lllllllllll}
  & & \multicolumn{4}{c}{\emph{xx}$\rightarrow$ \emph{en}} &  & \multicolumn{4}{c}{\emph{en}$\rightarrow$ \emph{xx}}\\
  \cline{3-6} \cline{8-11}
      & & \emph{Ar} & \emph{De} & \emph{He} & \emph{It} &  & \emph{Ar} & \emph{De} & \emph{He} & \emph{It} \\
     \multirow{2}{*}{(1)}&  \citet{aharoni-etal-2019-massively} & 27.84 & 30.50 & 34.37 & 33.64 &  & 12.95 & 23.31 & 23.66 & 30.33\\
     & \citet{philip-etal-2020-monolingual} & 32.99 & 37.36 & 39.00 & 39.73 &  & \textbf{17.22} & \textbf{29.94}  & \textbf{27.47} & 35.42\\
     & Our Bilingual
     & \textbf{33.11} & \textbf{39.01} & \textbf{39.11} & \textbf{41.40} & 
     & 16.79 & 29.73 & 26.80 & \textbf{36.23} \\
     \hline
     \multirow{5}{*}{(2)}& \citet{aharoni-etal-2019-massively} & 28.32 & 32.97 & 33.18 & 35.14 &  & 14.25 & 27.95 & 24.16 &  33.26\\
      & \citet{philip-etal-2020-monolingual} & 30.68 & 36.53 & 36.00 & 38.77 &  & 15.40 & 28.60 & 24.53 & 34.02\\
      & Parent {\sc mnmt} & 31.33 & 37.13 & 36.86 & 39.54 &  & 15.71 & 26.32 & 24.60 & 33.91\\
      & 50\% Pruned {\sc mnmt} & 30.84 & 37.10 & 36.29 & 39.44 &  & 15.41 & 26.20 & 24.06 & 33.70\\
      & Adapted {\sc mnmt} & 32.68 & 38.41 & 38.31 & 41.04 &  & 16.72 & 27.63 & 25.76 & 35.76\\
\end{tabular}
\caption{ {\sc bleu} scores of our models on the TED test sets compared to the literature, (1) - Bilingual baselines. (2) - Multilingual models scores. Here \citet{aharoni-etal-2019-massively} and \citet{philip-etal-2020-monolingual} are trained on 59 and 20 languages respectively. Parent {\sc mnmt} is our multilingual model trained till convergence on 9 languages. 50\% pruned {\sc mnmt} is the compressed parent {\sc mnmt}. Adapted {\sc mnmt} is the proposed model.}
\label{table:results}
\end{table*}

\section{Results and Discussions}
\subsection{Overcoming interference for high resource pairs:} 
In table \ref{table:results}, we present a comparative study of a high resource language scenario, severely affected by negative interference. Adapted {\sc mnmt} outperforms the parent {\sc mnmt} on all the 8 directions, with an average improvement of +1.40 on xx-en, and +1.32 on en-xx directions, and closes the gap with high performing bilingual baselines. 

\begin{figure}[ht]
\includegraphics[width=0.45\textwidth]{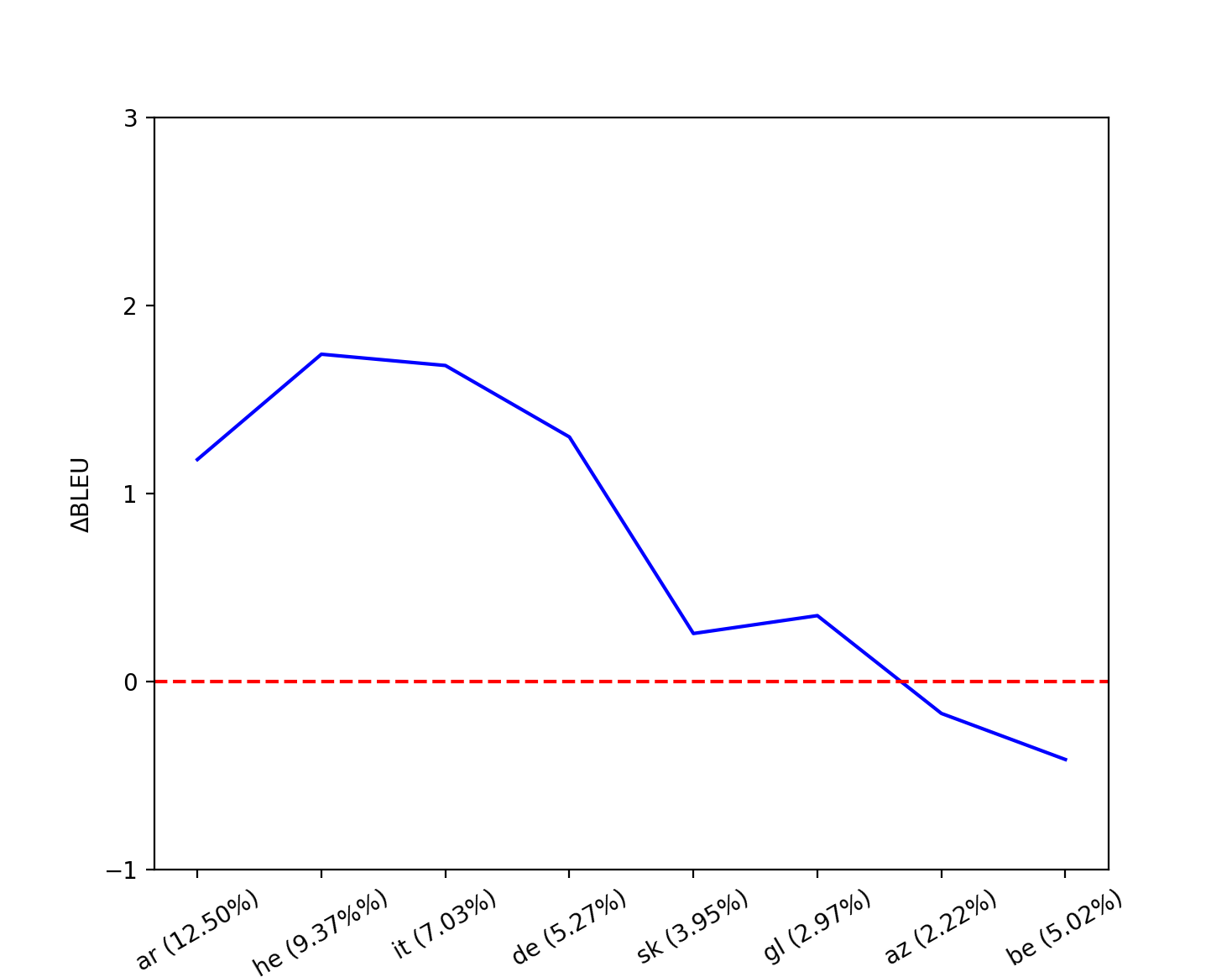}
\centering
\caption{Absolute difference in the BLEU scores, with the parent {\sc mnmt}, for 8 bilingual pairs. Each bilingual pair is the average over both the English-centric directions. The languages are arranged in the exact order of the training sequence. Numbers on the x-axis are percentages of the bilingual specific parameters used.}
\label{graph_1}
\end{figure}

\textbf{Analysing model capacity and negative interference:} Now, we expound on the problems of model capacity and interference. As shown in table \ref{table:results}, pruning 50\% of parameters from the parent {\sc mnmt} model leads to an average loss of just 0.29 {\sc bleu} points. This observation confirms, that there exists large redundancies even in a 9-language {\sc mnmt} model. The drop in the performance of an {\sc mnmt} over its counterpart bilingual models is loosely associated with the lack of capacity. As can be seen in figure \ref{graph_1}, by using only a fraction of parameters for each bilingual pair, we can significantly improve the performance over the parent {\sc mnmt}. Our results demonstrate the ability of parameter-free adaptation to fight negative interference, and improve the performance of severely affected high resource language pairs.

\begin{figure}[ht]
\includegraphics[width=0.45\textwidth]{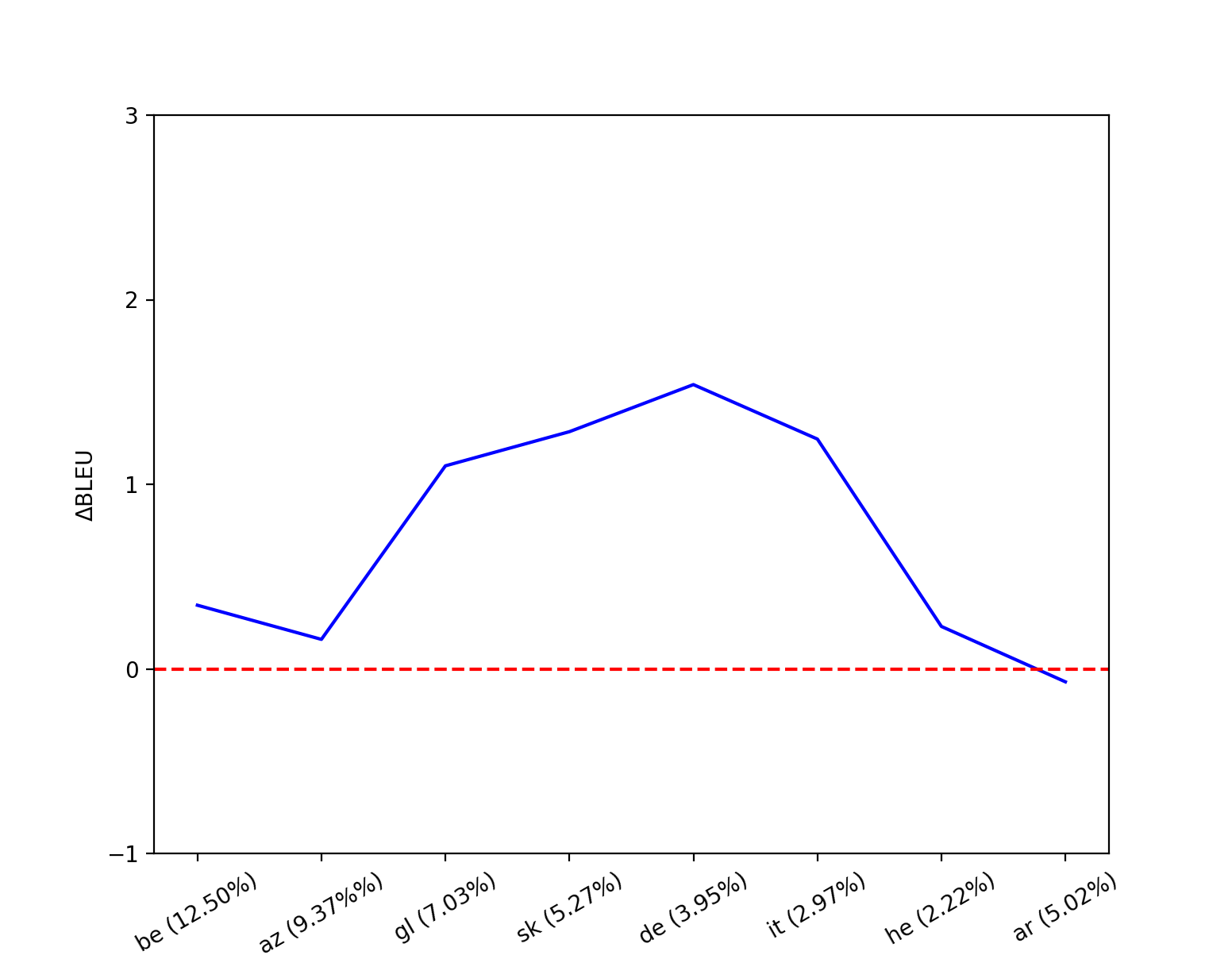}
\caption{Same as figure 2, trained in the reverse order}
\label{graph_2}
\centering
\end{figure}

\subsection{Analysing differences in the adaptation of high and low resource pairs:}
To understand the impact of parameter-free adaptation on both the high and low resource language pairs in an unbiased setting. We train two models in opposite orders of adding bilingual pairs. First, we train in the order of high to low resource languages (Ar to Be). Second, we train in the order of low to high resource languages (Be to Ar). Now, we assign the same proportion of parameters, to the high and low resource languages (Ar, He) in case 1, and (Be, Az) in case 2 respectively. As evident from figure \ref{graph_1} and \ref{graph_2}, the improvements in Ar and He in case 1 is significantly more, than the improvement in Be and Az in case 2. This observation agrees with the fact that negative interference severely affects the high resource languages in an {\sc mnmt}, and it needs adaptation to be improved. But, the performance of low resource languages in an {\sc mnmt}, is already near saturation due to the positive transfer from high resource languages. Hence, to extract the most out of parameter-free adaptation, it is better to prune and retrain the network in the order of high to low resource languages. This assigns high proportion of parameters to high resource pairs, to effectively overcome negative interference.

\begin{table*}
 \centering
 \begin{tabular}{lllllllllll}
  & & \multicolumn{4}{c}{\emph{xx}$\rightarrow$ \emph{en}} &  & \multicolumn{4}{c}{\emph{en}$\rightarrow$ \emph{xx}}\\
  \cline{3-6} \cline{8-11}
      & & \emph{Ar} & \emph{He} & \emph{It} & \emph{De} &  & \emph{Ar} & \emph{He} & \emph{It} & \emph{De} \\
     \multirow{1}{*}{(Bilingual)}&  Full-FT & 
     33.89 & 39.66 & 41.64 & 40.00 &  
     & 17.38 & 27.50 & 36.72 & 29.87\\
     \hline
     \multirow{5}{*}{(Multilingual)}& Ar only & 33.01 & - & - & - &  & 16.80 & - & - &  -\\
      & Ar-He & 33.21 & 38.26 & - & - &  & 16.72 & 25.52 & - & -\\
      & Ar-He-It & 32.99 & 38.45 & 41.03 & - &  & 16.61 & 26.00 & 35.56 - &\\
      & Ar-He-It-De & 32.68 & 38.43 & 41.14 & 38.39 &  & 16.72 & 25.89 & 36.08 & 27.25\\
\end{tabular}
\caption{ Full-FT represents the bilingual models derived from finetuning the full parent {\sc mnmt}. Rest are the adapted {\sc mnmt}s adapted over 50\% free parameters of the pruned {\sc mnmt}. 1) Ar only with 50\% parameters, 2) Ar, He with 25\% each, 3) Ar, He, It with 16.6\% each, and 4) Ar, He, It, De with 12.5\% each.}
\label{table:results_1}
\end{table*}

\subsection{Zero-shot Translation:}
\begin{figure}[ht]
\includegraphics[width=0.45\textwidth]{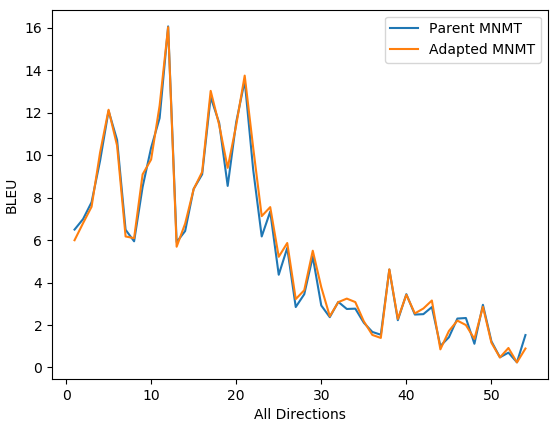}
\centering
\caption{Absolute BLEU scores for the parent and the adapted MNMT on all the 56 zero-shot xx-xx pairs arranged from high to low resource.}
\label{graph_3}
\end{figure}

Zero-shot translation in the context of {\sc mnmt}, refers to inference between pairs that are not seen directly during the training phase xx-xx. We show that we retain this important ability in our adapted {\sc mnmt}. Adapted {\sc mnmt} consists of 50\% pruned {\sc mnmt} weights and 50\% language specific weights. The pruned {\sc mnmt} weights are used to evaluate on zero-shot pairs, just like a traditional {\sc mnmt} by appending the source side language control token \citet{johnson-etal-2017-googles}. As shown in figure \ref{graph_3}, adapted {\sc mnmt} performs as good as the parent {\sc mnmt} on all the 56, xx-xx directions even with only 50\% of the total parameters. 

\subsection{Adapting to a subset of languages and retaining the multilinguality:}
Due to limited and fixed number of parameters, we cannot adapt to arbitrary number of languages. However, this framework allows high flexibility in adapting the parent {\sc mnmt} to only the languages of interest, while retaining the multilinguality simultaneously. We adapt the parent {\sc mnmt} to four models: 1) Ar, 2) Ar, He, 3) Ar, He, It and, 4) Ar, He, It, De. This way, we can assign all the free parameters to only the languages of interest and increase their capacities. The first pruning ratio is set to 50\% for all four models. The second pruning ratio is set such that each language receives equal proportion of parameters. From the results in table \ref{table:results_1}, we observe that assigning more parameters improve the performance marginally. The four adapted {\sc mnmt}s have similar performances, even with a significant difference in the proportion of parameters assigned for each language. The 4$^{th}$ model, with only 12.5\% parameters reserved for Ar, performs competitively with the 1$^{st}$ model with 50\% parameters for Ar. This implies, that a small fraction of parameters can effectively overcome negative interference, hence allowing space to adapt to multiple languages. To infer on the remaining languages which are not adapted, we can use 50\% pruned {\sc mnmt} weights, as done for zero-shot translation in the previous section, hence retaining the multilinguality.

In table \ref{table:results_1}, we also compare the results of the four adapted {\sc mnmt}s, with naive finetuning of the full parent {\sc mnmt} to bilingual pairs (Full-FT). The difference between naive finetuning and the proposed adaptation approach is that the former uses all the 100\% of model parameters and the embeddings to adapt to a single bilingual pair, thus the multilinguality is lost. While in our approach, the pruned {\sc mnmt} weights and the embeddings are fixed, and we only retrain the free parameters very efficiently, allowing to adapt to multiple languages. As can be seen in table \ref{table:results_1}, adapted {\sc mnmt}s perform competitively with Full-FT while retaining the multilinguality.

\section{Conclusion}
We investigate the problems of model capacity and negative interference in  multilingual neural machine translation. We show that even a 9 language {\sc mnmt} has a large proportion of redundant parameters, which are efficiently retrained to overcome interference. We propose a parameter-free adaptation strategy. Where, we use iterative pruning and retraining to improve bilingual representations, without any additional parameters. We hope that our work will attract more attention to practical and efficient ways of adapting an {\sc mnmt}.


\bibliographystyle{acl_natbib}
\bibliography{anthology,acl2021}


\end{document}